# Path planning model of mobile robot in the context of crowds

**W.Z. Wang, R.Q. Wang and G.H. Chen**

**Beijing University of Chemical Technology**

**Abstract**. Robot path planning model based on RNN and visual quality evaluation in the context of crowds is analyzed in this paper. Mobile robot path planning is the key to robot navigation and an important field in robot research. Let the motion space of the robot be a two-dimensional plane, and the motion of the robot is regarded as a kind of motion under the virtual artificial potential field force when the artificial potential field method is used for the path planning. Compared to simple image acquisition, image acquisition in a complex crowd environment requires image pre-processing first. We mainly use OpenCV calibration tools to pre-process the acquired images. In themethodology design, the RNN-based visual quality evaluation to filter background noise is conducted. After calibration, Gaussian noise and some other redundant information affecting the subsequent operations still exist in the image. Based on RNN, a new image quality evaluation algorithm is developed, and denoising is performed on this basis. Furthermore, the novel path planning model is designed and simulated. The expeirment compared with the state-of-the-art models have shown the robustness of the model.

## 1. Introduction

With the wide application field of mobile robots and the diversity of application environments in recent years, the path planning of mobile robots has become a hot research topic. Path planning is to guide the robot how to select the optimal route and control the robot to move from the starting point to the target point. The goal is to meet certain optimization indicators such as the least time spent, the shortest path, or the lowest energy consumption during the entire operation of the robot. Many mobile robot platforms have become more and more popular, and their applications have shifted from the static laboratory scenes to the highly dynamic environments with complex interactions between humans and robots while making it more difficult to find suitable paths and may need to deal with moving objects around, so path planning for mobile robots is particularly important [1, 2, 3].

Path planning is the process of automatically generating feasible and optimal paths to a predetermined target point under static and dynamic environments, model constraints and uncertainties. Based on the literature review, the state-of-the-art models can be summarized as the follows [4, 5]. (1) The basic idea of the artificial potential field method is based on the virtual force method to simulate the force analysis of the robot moving in the active environment. This method has a simple structure, but the robot is likely to be trapped under the action of the artificially abstract field at a certain balance point, a deadlock phenomenon occurs, which may cause the mobile robot to stay at a certain local balance point before reaching the target point, thereby causing a deadlock. (2) The grid decomposition method uses the same size squares to divide the environment that the robot may search for, and the partitioned squares are replaced by arrays in the program. The divided environment is divided into two categories, the robot can move freely in the area, blocking the robot movement of the roadblock area [6, 7, 8, 9]. (3) The planning algorithm based on intelligent bionics can solve the problem by studying some social behaviors in nature and summing up the motion laws according to their principles, and finally imitating the structure and behavior. Such as ant colony algorithm, genetic algorithm, particle swarm algorithm, neural network, etc., currently mostly used for the robot coordination control, power system optimization, network routing optimization [10, 11, 12]. (4) SBP algorithm based on sampling planning is the most popular and influential path planning algorithm, which is simple, general, efficient and probabilistic. It can select a finite number of non-structural points to connect with each other and construct a road map of feasible trajectories without clear information of obstacles. (5) Global path planning is to obtain the optimal path from the initial place to the target place under the known environment information and in the pre-built environment model. Its advantage is to ensure the optimization and accessibility of the planned path. Local path planning is to obtain the information of the surrounding environment based on the sensor in the unknown environment, and to then make the robot automatically obtain an optimal path without collision. Its advantages are more adaptability to the unknown environment and real-time planning. (6) Local path planning is to obtain the surrounding environment information based on sensors in an unknown environment, and make the robot autonomously obtain an optimal path without collision. Classical algorithms for the local path planning include artificial potential field method, simulated annealing method, fuzzy logic method, neural network method, dynamic window method, reinforcement learning method, and behavior-based path planning method [13, 14].

In the past research, the models are more focused on the planning without the sufficient studies on the vision information. Inspired by the mage quality assessment and the OpenCV framework, this paper proposes the planning model based on RNN and visual quality evaluation in the context of crowds. The rest of the paper is organized as the follows. In the setion 2, we introduce the image collection in the background of crowd complex environment. Compared to simple image acquisition, image acquisition in a complex crowd environment requires image pre-processing first. This part mainly uses OpenCV calibration tools to pre-process the acquired images. In the section 3, the RNN-based visual quality evaluation to filter background noise is conducted. After calibration, Gaussian noise and some other redundant information affecting the subsequent operations still exist in the image. Based on

RNN, a new image quality evaluation algorithm is developed, and denoising is performed on this basis. In the section 4, scene recognition for robot path planning is studied. RNN is still used for scene recognition. After recognizing the scene, neural network is used for path planning based on interaction, and comparative experiments are carried out at the same time. The section 5 gives the finalized experiment compared with the other methodologies. In the section 6, the conclusion is done and the prospect is provided for guiding the further research.

## 2. Image Collection and Pre-processing

The crowd state perception process mainly includes the four major processes, namely video acquisition and processing, optical flow calculation, vector optical flow field mapping to geographic space, and the crowd state detection and analysis. The purpose of the behavior analysis method is to make the computer understand the human behavior through the analysis of the crowd behavior in the image, so as to realize intelligent monitoring. Among them, feature extraction and classification recognition of crowd behavior are the core content of behavior analysis algorithms. Traditional behavior analysis methods mainly include analysis methods based on behavior feature extraction, behavior analysis methods based on wavelet transform algorithm and analysis methods based on rigid body kinematics, of which the most commonly used are behavior analysis methods based on behavior feature extraction.

In our pre-processing model, the listed aspects are considered [15, 16]. (1) Based on the high-definition remote sensing images or GPS, collect the position of the camera's center point. Based on camera calibration or direct setting, obtain the internal and external reference information such as the camera's height, tilt, azimuth, main distance, etc. (2) The optical flow field of each surveillance video is calculated. Based on the two images in the video, the relevant optical flow calculation method is adopted to obtain the real-time flow field of the current video. (3) In the geospatial space, any vector optical flow has a clear spatial reference, it is easy to understand the position, direction, speed, acceleration and other motion parameters of the crowd, which is an important basis for crowd state detection and early warning.

The development of the system involves the multiple hardware platforms and multiple software development frameworks. The hardware platform includes the PC terminal and the embedded development board. How to develop applications on different platforms, how to port the final applications on different platforms, and how to communicate between different platforms are all issues that need to be considered. Software development framework mainly includes the Qt application development framework, OpenCV computer vision library, PyTorch deep learning framework, how to make full use of the characteristics of each open source framework, how to develop collaboratively under different frameworks, these need to constantly learn and explore in the development process. Inspired by this, in the process of using the traditional behavior analysis method to detect the abnormal behavior of the crowd in the sensitive area, it is inevitable that the image collection equipment will reduce the human behavior characteristics. To this end, a method for detecting abnormal behavior of crowds in sensitive areas based on the weighted Euclidean distance behavior analysis is proposed. This article uses the Lucas-Kanade optical flow method to obtain the movement state of the crowd. This method first converts the video into a sequence image and converts it into a sequence grayscale image. Constrained in a short time, the gray value of the same point of the moving target will not When there is a change, the equation is constructed accordingly, and the optical flow vector is calculated [17, 18, 19].

There are various points to be extracted in the image sequence and the other background noise. To extract the target points from the multi-valued digital image, it must be binarized. The feature points are presented as the formula 1.

$$I(u,v) = \begin{cases} 255, & I_{Calculation}(u,v) > T \\ 0, & I_{Calculation}(u,v) < T \end{cases} \quad (1)$$

Where the $I(u,v)$ represent the pixel value of a point in the image. In the process of using the behavior analysis method to detect the abnormal behavior of the crowd in the sensitive area, it is necessary to construct a characteristic database of the abnormal behavior of the crowd, and use the characteristic database to intelligently detect the abnormal behavior of the crowd, the estimation is presented as follows.

$$y_{estimation}(n) = -\sum_{i=1}^{k} a_i \times y(n-i), n \in N \quad (2)$$

Where the $N$ is the natural number, the $a_i$ is the weights of the behavioral characteristics. The above formula can be described by the mean square error criterion as the formula 3.

$$\ell_{min} = E[e^2(n)] \quad (3)$$

Different behaviors of crowds are manifested in the different forms, and the measurement standards of behavior characteristics are also different, which will cause the problem that different behavior characteristics are masked. Therefore, it is necessary to standardize the attributes of behavior characteristics. Then, for further processing, the OpenCV is adopted. The system hardware consists of host, video capture camera, network device and storage medium. Based on the software architecture on the host computer, the video image data in the monitoring area is collected by camera, and the image preprocessing is carried out. Then the relevant code is written by the OpenCV visual processing library to detect the collected video image data, and the moving object is detected and judged in the threshold range. Finally, the alarm mechanism is triggered. OpenCV is an image processing function library, which contains API interfaces and the source code libraries supported for computer operating systems and computer language programming platforms. We can judge the content of images with your own program code to achieve recognition operations. The image is processed by calculating the approximate gradient of the image gray function, so this algorithm is used to complete the optimization of image detection. Because the system needs to process in real time, it must have high requirements for processing speed. Based on FPGA, it can realize the advantages of parallel processing of signals and abundant logic resources. Therefore, FPGA is used as a carrier for image processing.

**3. RNN-based Visual Quality Evaluation and Noise Removal**

With the development of the digital image technology, image quality evaluation plays an important role in image processing such as image compression, transmission, watermarking and pattern recognition, so it is of great significance to evaluate image quality effectively. Man is the ultimate sensation of the image, and the quality of the image is judged mainly by whether it is better in line with the human perception. Therefore, an image quality evaluation algorithm that conforms to the perceptual characteristics of the human eye should make full use of the visual characteristics of the human eye. Due to the insufficient use of the structural similarity evaluation method in terms of the visual characteristics, the gradient amplitude containing important information for understanding the scene is considered, but there are also some problems in using the gradient amplitude alone to improve the structural similarity evaluation method, so the different the contrast sensitivity characteristics of the human eye for the difference in frequency perception are further improved. Inspired by the previous research, the features are summarized as follows [20].

(1) The multi-channel theory of vision believes that the human visual system is a Fourier analyzer, where there are many channels, and the spatial frequency region modulated by each channel is different. Although the expression of CSF function obtained by different experiments is different, it is generally considered that the contrast sensitive characteristic of human eye is a function of the spatial frequency, which has certain properties of bandpass filter and different directions.

(2) Spatial frequency refers to the gray spatial change rate within the image area. Experiments show that the visibility of human vision to information changes with the change of local spatial frequency in the image area, which is called the visual spatial frequency masking effect. The masking effect consists of the contrast masking and entropy masking: contrast masking takes into account the contrast correction of the visual threshold and entropy masking takes into account the effect of the local neighborhood features on the vision. If the new observation set can be used to completely reconstruct the signal, the original observation set is updated with the new observation set and the above process is repeated; otherwise, the iteration ends and the result of the previous round is used as the observation set of the next stage. Then, in the observation set obtained in the previous stage, each observation is divided into key observations and non-key observations based on the importance of the signal reconstruction, and some non-key observations are eliminated under the premise of ensuring complete signal reconstruction.

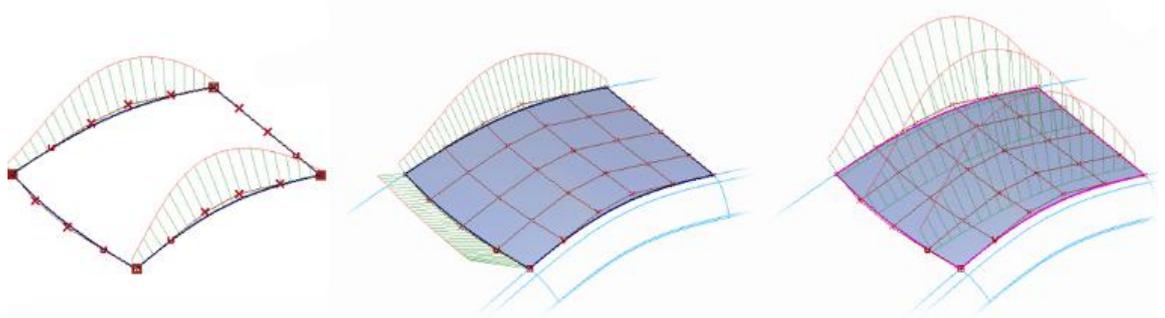

**Figure 1. The Visual Quality Curve**

In the figure 1, we show the visual quality curve for the references. We take the reference [11] to construct the quality assessment level as the follows.

$$\begin{cases} f_x = \sqrt{\dfrac{1}{MN} \sum_{i=1}^{N} \sum_{j=2}^{N} [I(i,j) - I(i, j-1)]^2} \\ f_y = \sqrt{\dfrac{1}{MN} \sum_{i=2}^{N} \sum_{j=1}^{N} [I(i,j) - I(i-1, j)]^2} \end{cases} \quad (4)$$

The ability of the human visual system to resolve details is related to the relative contrast of the area viewed, usually expressed by the contrast sensitivity function CSF. For this idea, the RNN will assist the noise removal task. The traditional neural network consists of a series of simple neurons. Simple NN includes input layer, hidden layer and output layer. The hierarchical structure is usually that each layer of neurons is fully connected to the next layer of neurons, and there is no connection between neurons of the same layer. There is no loop or loop in the network structure, and there is no feedback connection between the output of the network and the model itself. In NN, all observations are processed independently of each other [21].

However, the data in many tasks is rich in a lot of context information, and there is also a complex correlation between each other. RNN is mainly used for processing sequence data. Its biggest feature is that the output of the neuron at a certain time can be input to the neuron again as an input. This kind of network structure in series is very suitable for time series data and can maintain the dependency relationship in the data. In the figure 2, we present the original RNN framework.

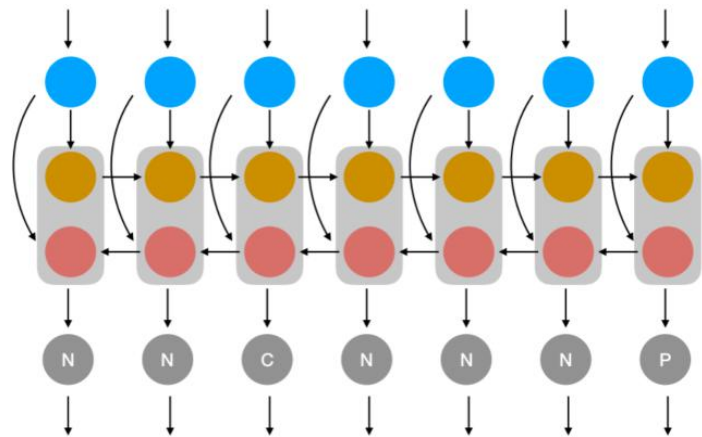

**Figure 2. The Original RNN Framework**

The image repair method based on deep learning is similar to using the deep learning technology to solve other problems. Using the deep self-encoder network structure, for different input information, different output information can be set for training, so that the trained network establishes the corresponding relationship between the input information and the output information. Therefore, the network training becomes an issue that must be considered. For the designed methodology, we are focused on following aspects. (1) A fast multi-scale residual block is proposed as an important component of the network, which also makes the network more adaptable to the fuzzy inputs of some different scales. (2) A novel multiscale deep convolutional network is proposed for blind deblurring of heterogeneous images. The network uses a multi-scale structure from thick to thin to gradually restore clear images, and the same parameter settings are used for each scale. Parameter sharing among different scales can not only reduce the number of the parameters, but also prevent overfitting. At the same time, the structure of codec and jump connection is used in all scales, which can accelerate the convergence of the network and restore the texture information of the image better.

## 4. Robot Path Planning Model in the Context of Crowds

At present, the traditional reinforcement learning method is used in the path planning of mobile robots. Because it has no prior knowledge of the environment at the initial stage of learning, it often has problems such as slow convergence speed and long learning time. Based on the studies in the previous sections, the collected visual information is pre-processed with the registration and noise removal, which will enhance the accuracy of capturing the information. Based on the above, the current multi-robot system task allocation needs to solve the following problems: 1) There are no obstacles in the two-dimensional workspace, so the multi-robot task allocation algorithm does not have practical benefits; 2) The direction angle of the robot movement is not considered In the actual environment, the original RNN algorithm directly uses the deviation of the robot's current position and the target position in the environment to plan the path of the robot.This may cause obstacles to cross, so that the robot cannot reach the corresponding task target point so that the task assignment cannot be completed successfully. In this paper, we propose a biologically inspired method to generate a working map of the robot. Biologically inspired methods include several sub-steps [22, 23].

Generally, the biological heuristic method first randomly generates a core node through collision detection in the work map. The process of collision detection is to determine whether a randomly generated node is a feasible node, and a node that is on the map and does not conflict with an obstacle is a feasible node. Inspired by the models in the biological nervous system, the literature [7] proposed a biologically inspired neural network method for solving the path planning problem of mobile robots, with the status function as formula 5, where the $[I_i]^+ + \sum_{j=1}^{k} w_{ij}[x_i]^+$ denotes the irritation and inhibitory input of neurons.

$$\frac{dx_i}{dt} = -Ax_i + (B - x_i)\left([I_i]^+ + \sum_{j=1}^{k} w_{ij}[x_i]^+\right) - (D + x_i)[I_i]^- \quad (5)$$

This neural network method does not require a learning process. According to the information transfer between neuronal cells, the potential function of the state of the neuronal cells can be obtained. For a single robot, the cost is measured by the distance from the starting point to all the target points it passes, while for the entire multi-robot system, the cost is the sum of the costs of all single robots. In order to complete access to all the targets in the work area, the first is to examine the path planning. Because there are obstacles in the work area, each robot must have the ability to avoid obstacles. For this requirememnt, the Q-learning and the dynamic model should be integrated. When the winning neuron is selected, the next step is to design the neighborhood function, and then determine the next winning neuron. The function of neighborhood function is to determine the influence between winning neuron and the neighborhood neuron. The influence of the winning neuron is that the greater the intensity, the lower the influence of the neuron adjacent to the winning neuron until it has no effect. If there is no effect, it shows that the neuron is no longer neighborhood neuron of winning neuron, that is, the non-neighborhood neuron. In the working area, the neighborhood of the winning neuron is a circle, and the coordinate of its center is the position coordinate of the winning neuron. With the process of updating the iteration, the center of the circle changes all the time until the winning neuron reaches the corresponding task target point. In the formula 6~7, we model the dynbamic process.

$$x_i(t) = \begin{cases} -Ax_i(t) + D_i(T)y_i(t), & scene1 \\ -Ax(t) + D_i(T)y_i(t) + I, & scene2 \end{cases} \quad (6)$$

$$y_i(t) = g\left(\sum_{j \in NE_i} w_{ij} x_j(t)\right) \quad (7)$$

After the winning neuron and its winning neighbors are determined, the winning neuron and its neighboring neurons need to move to input neuron corresponding to the winning neuron. In this paper, we extend a neural network model for real-time mobile path planning of mobile robots in RNN, where the optimal path is generated by distributed consensus. Bias minimum consensus protocol, the word bias in this protocol is reflected in considering the influence of the distance between two nodes during network communication. Now, the information the node receives from its neighbors is $\{s_q + D_{pq}\}$, since the following node receives information from its neighbors and the leading node does not receive any information from its neighbors, the bias minimal consensus protocol is distributed in the minimum consensus protocol, we know that the leader node is static, with a time derivative of zero, and does not receive any information from its neighbor node. In addition, the following nodes are dynamic, and they receive information from neighbor nodes during communication, because the communication delay caused by the distance between two nodes receiving information is actually $\{s_q + D_{pq}\}$, the model should have the following restriction condition.

$$\begin{cases} s_p = 0, & p \in N_1 \\ \delta s_p = -s_p + \min_{q \in N(p)} \{s_q + D_{pq}\}, & p \in N_2 \end{cases} \quad (8)$$

$-s_p + \min_{q \in N(p)} \{s_q + D_{pq}\}$ means that change of the random node state value is the result of the feedback between their state value and the information received from the neighbors. The feedback strength can be adjusted through the change of the parameters. In the figue 3 and 4, we present the biologically inspired method illustration and enhanced neural network framework, respectively.

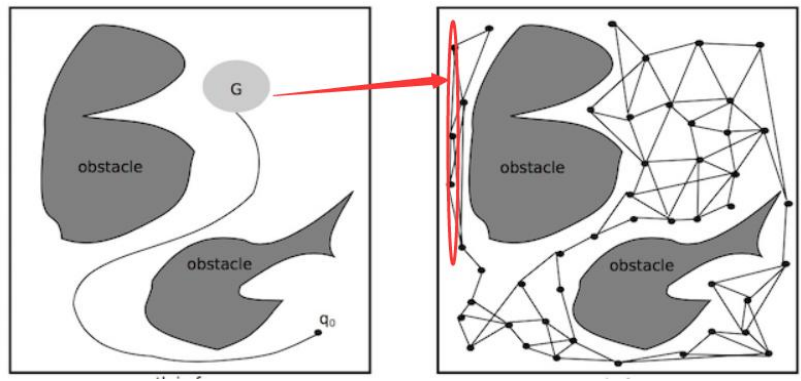

Figure 3. The Biologically Inspired Method Illustration

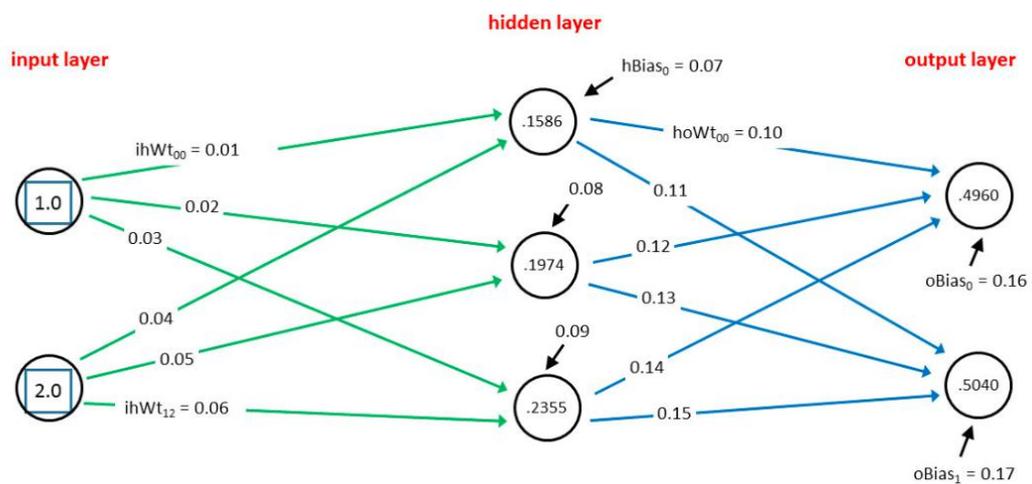

Figure 4. The Enhanced Neural Network Framework

The modeling of mobile robot working environment is to abstract the real three-dimensional space environment information into a three-dimensional mathematical model. In this paper, the three-dimensional space environment model is a topology of the basic idea of the two-dimensional grid method to three-dimensional space, and the working environment of the robot is discretized in units of grids. The static path planning simulation simulation robot is known to the environment, and the starting point, the target point, and the obstacle point are all known.It is necessary to implement the planning of the simulated robot to find the optimal path from the starting point to the target point in the algorithm. Simulation environment is on the computer to imitate the real environment, designers will actually abstracted into virtual environment, and the virtual environment with algorithmic language in the form of the chart image or image, etc. According to the theoretical basis, we have to simulate the robot activities of environment should be the starting point and end point, there is a fixed obstacles, through the array to the environment simulated in an array have to be marked on every point of the simulation environment, and attach the corresponding meaning.

In this paper, the target point is moving at a non-uniform speed. If only the relative position is considered, this may cause the robot to only reach the target point, but it cannot maintain the same movement trend as the target point. So for question one, the relative speed and relative acceleration between the robot and the target point are introduced in the article, the formula 9 defines the framework.

$$F(X,V,a) = \alpha(X_g - X) + \beta(V_g - V) + \lambda(a_g - a) \qquad (9)$$

According to the above formula of gravitational function, the gravitational force is 0 only the relative position between the robot and the target point, the relative velocity and the relative acceleration are both zero. This can not only ensure that the robot reaches the target point, but also ensure that the robot can track the target point in real time. For optimizing the model, the GA and enhanced PSO will be integrated. The selection operator compares the individual fitness value f with the random number R, and retains the individuals whose fitness value is greater than the random number R, and discards the rest. This selection operator keeps the path of individuals always developing in the optimal direction, and selection by comparing with random number R not only maintains individual diversity, but also ensures individual optimality. According to the idea of biological chromosome recombination, the crossover operator divides the initial population M individual randomly into the chromosomes with equal length, then exchanges the two segments in the corresponding chromosome position to generate new path individuals.

The mutation operation is to increase the diversity of the individual by changing the chromosomal genes in the individual. Different encoding methods have different mutation methods. For the binary encoding method, the mutation is the change of 0 → 1 and 1 → 0 encoded by the individual. For the encoding method according to the serial number, the method of mutation is usually to randomly select two discontinuous gene fragments, and then search for the starting and ending positions according to the shortest path principle and the non-repetitive principle to form a complete individual. The above improved repulsion function formula is applicable to any position and position of the obstacle at the target point, but when a special force situation occurs, the robot may still collide with the obstacle, that is: one of the obstacles is located in the robot between the target point and the movement direction of the three, the other obstacles are randomly distributed. When the combined force of gravity and repulsion of the robot can make the robot continue to advance in the current direction, if the movement state of the target point and all obstacles has remained unchanged, the robot will continue to advance under the effect of the combined force and gradually approach the obstacle At this time, the robot is likely to collide with obstacles, causing unnecessary losses. Hence, in the formula 10, we set the revised fitness function.

$$fitness = 1/d\left(1 + \frac{1}{\sqrt{n-1}} + R\right) \qquad (10)$$

Find from the target point from the back to the front, each time determine the point in the four neighborhoods of the current point that is closest to the current g value, and then set the point as the current point and set the point The p value of the point (to judge whether it is a point in the optimal path) is 0, and it is extrapolated by this example until the starting point is found.

After optimization, a relatively complicated pathfinding map is obtained. The ant colony algorithm updates the pheromone every iteration in order to then avoid accumulating the pheromone left behind and overwhelm the heuristic information. However, the traditional pheromone update formula performs the same treatment for the pheromone left by the excellent ants and the inferior ants. In this way, the guidance of the optimal solution of each generation cannot be fully displayed, and the pheromone of inferior ants is also equivalent to a disturbance, which will reduce the efficiency of the ant colony. In the formula 11~12, we update the regulation functions.

$$\delta_{ij}(t+1) = (1-\beta)\delta_{ij}(t) + \Delta\delta_{ij}(t,t+1) \qquad (11)$$

$$\Delta\delta_{ij}(t,t+1) = \sum_{k=1}^{m}\Delta\delta^{k}_{ij}(t,t+1) \qquad (12)$$

In order to prevent the robot from colliding with obstacles before taking emergency obstacle avoidance measures in actual situations, a method of judging collision is used in this paper, that is, given a safe distance, when the robot and the obstacle When the distance is less than the safe distance, it can be considered that the robot collides with the obstacle, at this time the robot will take emergency obstacle avoidance measures. For this measure, the BSO model should be then considered. The basic principle of the algorithm is to replace the particles in the particle swarm optimization algorithm with the beetle, that is, to use BAS optimization to replace the individual optimal value comparison in the particle swarm optimization algorithm. The initial position and velocity parameters of the beetle are the same as those in the particle swarm optimization algorithm. In the iterative process, the way to update the position of the longicorn depends not only on the current global optimal solution of the longicorn individual, but in each iteration, the longicorn judges the odor concentration. Individuals in the beetle herd will update the position of the beetle herd in the iterative process by comparing the fitness function values of the odor concentrations on the left and right sides of the beetle herd. The beetle swarm algorithm constructed by this method can effectively overcome the problems of step size reduction in the BAS algorithm resulting in local optimization and poor stability of the particle swarm algorithm.

Since the movement direction of the beetle is randomly selected, the beetle direction random vector is established and normalized as follows.

$$b_{movement} = \frac{rand(n,1)}{\|rand(n,1)\|} \quad (13)$$

In the chemotaxis operator of the algorithm, bacteria have the ability of the self-judgment and automatic adjustment of the target space perception. In order to improve the individual's ability to find the best, the search method will be divided according to the size of its fitness value: if it is better than 80% of the fitness value, the best bacterial individual will be tracked; if it is worse than 80%, the tracking bacteria search at the center of the group; if both methods fail, then search randomly; if the random search fails, turn the search direction 180° to ensure that the algorithm has a certain convergence speed. In the figure 5, we present the designed searching framework. In the next section, we will conduct the research of simulations to validate the overall performance.

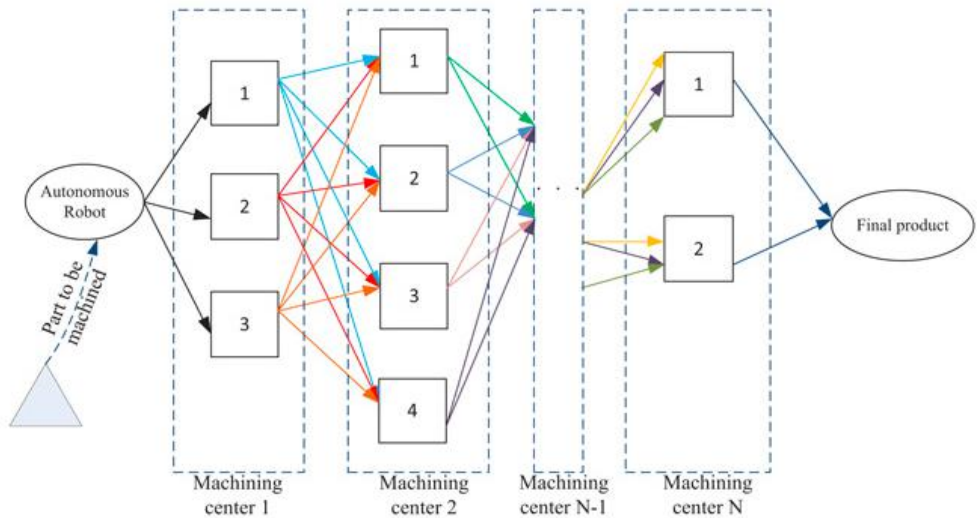

Figure 5. The Designed Searching Framework for the References

## 5. Experiment and Verifications

This article uses the simulation system of dynamic path planning. Using Google's open source TensorFlow module to build a deep reinforcement learning DDPG network architecture, using the OpenAI Gym environment library to build a free floating space robot dynamic path planning simulation system.

In the MATLAB simulation environment, experiments were conducted on the path planning of the artificial potential field method before and after the improvement. In the experiment, the robot was reduced to a particle, the obstacle was reduced to a circle, and the target was reduced to a triangle. To begin with, in the figure 6, we present the experiment environment. In the training process, the number of rounds set in this scene is 3000, the maximum training step for each round is 300 steps, the termination condition of each round of training is that the robot reaches the target point or reaches the maximum number of steps, and the training time is 12 hours. In the figure 7, we present the experiment on the simulation efficiency. It can be seen from the number of iterations that the two algorithms find the optimal path, the BSO algorithm is significantly stronger than the ACO algorithm in adaptability to the environment and convergence performance. From the running time of the two algorithms, it can be seen that the average running time of the BSO algorithm under different path start and end points is only 40.73% of the ACO algorithm. In the figure 8, we present he path searching result. As the number of training increases, the reward obtained by the space robot gradually becomes larger, indicating that the robot is moving closer to the optimal path, and eventually converges to around 0.9, indicating that the robot has found the optimal path. Because the target point varies randomly, the reward is not a fixed value , The reason for the occurrence of "noise" is consistent with scenario 1. In the figure 9, the finalized searching and path planning scenario is simulated.

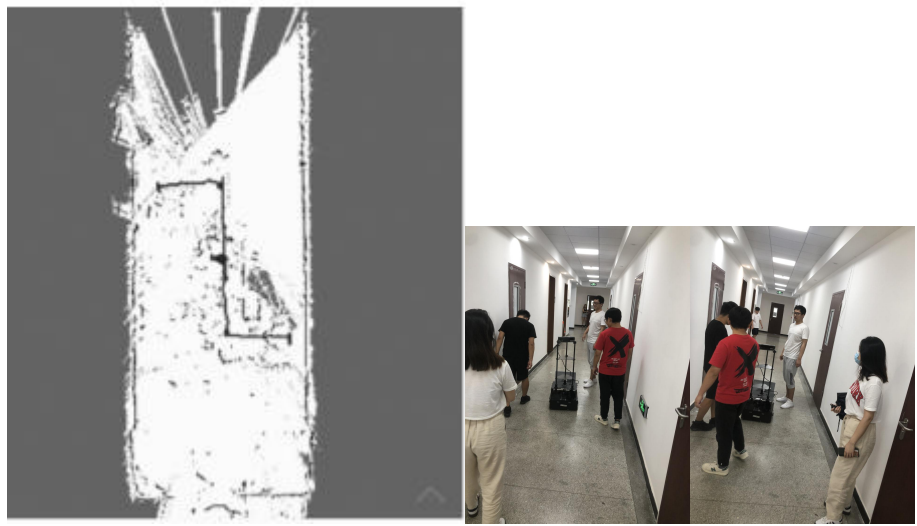

**Figure 6. The Experiment Environment**

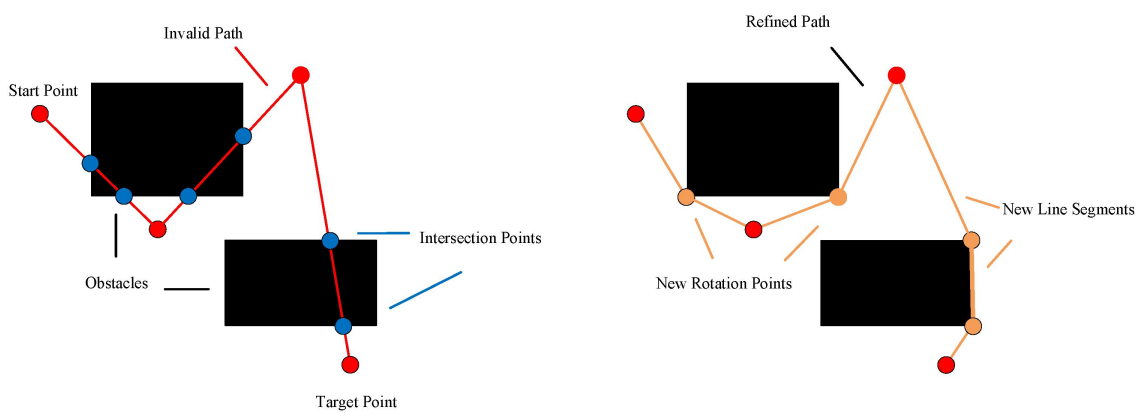

**Figure 7. The Experiment on the Updating Efficiency**

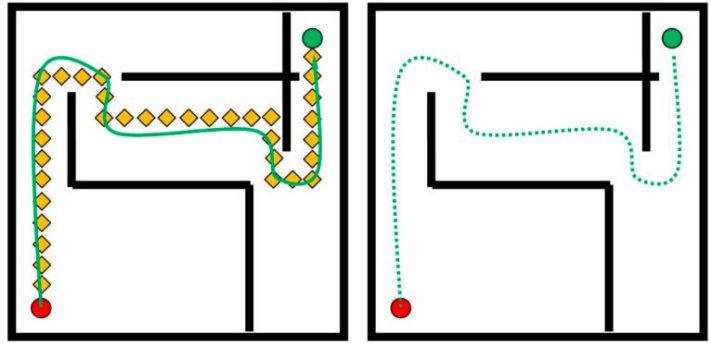

**Figure 8. The Path Searching Simulation Result**

![Figure 9 grid]

**Figure 9. The Finalized Path Planning Experiment**

## 6. Conclusions

Robot path planning model based on RNN and visual quality evaluation in the context of crowds is analyzed in this paper. It aims to enrich path planning methods and provide an effective choice for mobile robots to plan paths in a specific environment. The algorithm has the advantages of simple modeling, less adjustment parameters, less calculation, and fast convergence speed. Compared to simple image acquisition, image acquisition in a complex crowd environment requires image pre-processing first. We mainly use OpenCV calibration tools to pre-process the acquired images. In themethodology design, the RNN-based visual quality evaluation to filter background noise is conducted. After calibration, Gaussian noise and some other redundant information affecting the subsequent operations still exist in the image. Based on RNN, a new image quality evaluation algorithm is developed, and denoising is performed on this basis. Furthermore, the novel path planning model is designed and simulated. In the future research, we will consider the hardware implementation and testing on the proposed model.